\documentclass[letterpaper]{article} 
\usepackage{aaai2026}  
\usepackage{times}  
\usepackage{helvet}  
\usepackage{courier}  
\usepackage[hyphens]{url}  
\usepackage{graphicx} 
\usepackage{natbib}  
\usepackage{caption} 
\usepackage{amsmath}
\usepackage{amsfonts}
\usepackage{multirow}
\usepackage{algorithm}
\usepackage{algorithmic}

\usepackage{newfloat}
\usepackage{listings}
\DeclareCaptionStyle{ruled}{labelfont=normalfont,labelsep=colon,strut=off} 
\floatstyle{ruled}
\newfloat{listing}{tb}{lst}{}
\floatname{listing}{Listing}
\title{Deconstructing Pre-training: Knowledge Attribution Analysis in MoE and Dense
Models}
\author {
    Bo Wang\textsuperscript{\rm 1}, 
    Junzhuo Li\textsuperscript{\rm 1,\rm 2},
    Hong Chen\textsuperscript{\rm 1},
    Yuanlin Chu\textsuperscript{\rm 1},
    Yuxuan Fan\textsuperscript{\rm 1},
    Xuming Hu\textsuperscript{\rm 1,\rm 2}\thanks{Corresponding author}
}
\affiliations {
    \textsuperscript{\rm 1}The Hong Kong University of Science and Technology (Guangzhou)\\
    \textsuperscript{\rm 2}The Hong Kong University of Science and Technology\\
    \{bo.wang, jz.li, hchen763, ychu763, yfan546\}@connect.hkust-gz.edu.cn\\
    xuminghu@hkust-gz.edu.cn
}

\begin{document}

\maketitle

\begin{abstract}
Mixture-of-Experts (MoE) architectures decouple model capacity from per-token computation, enabling scaling beyond the computational limits imposed by dense scaling laws. Yet how MoE architectures shape knowledge acquisition during pre-training—and how this process differs from dense architectures—remains unknown. To address this issue, we introduce Gated-LPI (Log-Probability Increase), a neuron-level attribution metric that decomposes log-probability increase across neurons. We present a time-resolved comparison of knowledge acquisition dynamics in MoE and dense architectures, tracking checkpoints over 1.2M ($\approx 5.0\text{T}$ tokens) and 600K ($\approx 2.5\text{T}$ tokens) training steps, respectively. Our experiments uncover three patterns: (1) \textbf{Low-entropy backbone.} The top approximately 1\% of MoE neurons capture over 45\% of positive updates, forming a high-utility core, which is absent in the dense baseline. (2) \textbf{Early consolidation.} The MoE model locks into a stable importance profile within $<$ 100K steps, whereas the dense model remains volatile throughout training. (3) \textbf{Functional robustness.} Masking the ten most important MoE attention heads reduces relational HIT@10 by $<$ 10\%, compared with $>$50\% for the dense model, showing that sparsity fosters distributed---rather than brittle---knowledge storage. These patterns collectively demonstrate that sparsity fosters an intrinsically stable and distributed computational backbone from early in training, helping bridge the gap between sparse architectures and training-time interpretability.
\end{abstract}


\section{Introduction}

Large language models (LLMs) continue to scale in parameter count, but dense architectures make such scaling computationally inefficient, given the near-linear growth of per-token FLOPs with model size~\citep{kaplan2020scaling,hoffmann2022training}. To mitigate this, Mixture-of-Experts (MoE) architectures leverage conditional computation, expanding total capacity while activating only a sparse subset of experts per token, thus decoupling parameter growth from per-token compute~\citep{shazeer2017outrageously,lepikhin2020gshard,abnar2025parameters}. Building on this idea, recent large-scale LLMs increasingly adopt MoE architectures~\citep{liu2024deepseek,comanici2025gemini,cai2025survey,mu2025comprehensive}, achieving strong downstream performance while scaling to trillion-parameter regimes under manageable per-token compute~\citep{yin2025pangu,team2025kimi,yang2025qwen3,li2025every}.

Despite the impressive performance and efficiency of MoE models, our understanding of their internal working mechanisms remains limited. Recent interpretability research has developed a suite of neuron attribution techniques—such as Integrated Gradients (IG)~\citep{sundararajan2017axiomatic,dai2021knowledge}, Layer-wise Relevance Propagation (LRP)~\citep{chefer2021transformer,ali2022xai}, Log-Probability Increase (LPI)~\citep{yu2023neuron,yu2024interpreting}, and Neuron-Grad (NEG)~\citep{zhao-etal-2025-neuron}. However, these studies are constrained by two key limitations. First, they have focused on traditional dense Transformers. Second, existing analyses are almost post-hoc, dissecting models only after they have been fully trained. Recent studies have begun to trace representation during learning—for instance, linear probes that chart the rise of trustworthiness dimensions across 360 checkpoints, and circuit-tracing work that follows knowledge sub-graphs through finetuning~\citep{qian2024towards,ou-etal-2025-llms}. However, these efforts again focus on dense models, leaving the question: how architectural differences influence the dynamic knowledge acquisition process.

In this paper, we present a neuron-level, time-resolved comparison of MoE and dense Transformers throughout pre-training to fill this gap. Building on LPI, we introduce a gated attribution measure that decomposes log-probability increase across experts (feed-forward network), and attention heads. We apply this method to OLMoE-1B-7B (MoE) and OLMo-7B (dense), two open-source Transformers, with 1.2M training steps ($\approx 5.0\text{T}$ tokens) and 600K steps ($\approx 2.5\text{T}$ tokens), respectively. By tracking pre-training checkpoints, we quantify and compare the evolution of importance for individual neurons in both the feed-forward network (FFN) and attention (ATTN) layers, and analyze how their importance evolves from three complementary perspectives: (i) neuron-level stability, (ii) layer-level consolidation dynamics, and (iii) functional robustness. We uncover three findings:
\begin{itemize}
\item \textbf{Low-entropy MoE backbone.} Top approximately 1\% of MoE neurons receive $>$ 45\% of positive updates, and this set is preserved---and reinforced---throughout training, contrasting with the churn observed in the dense model.

\item \textbf{Early consolidation in MoE.} Less than 100K steps suffice for MoE FFN and ATTN layers to lock into a stable importance profile, whereas the dense model exhibits volatility.

\item \textbf{Functional robustness.} Ablating the ten most important MoE attention heads reduces relational HIT@10 by $<\!10\%$, and the top $1\%$ of MoE FFN neurons by $\approx35\%$, whereas the dense baseline suffers $\approx50\%$ and $\approx96\%$ drops, respectively---indicating that sparsity fosters distributed rather than brittle knowledge storage.
\end{itemize}

By coupling scaling-friendly architectures with pre-training interpretability, our study bridges two previously disjoint threads and reveals that MoE models rapidly consolidate a stable and distributed backbone, from neurons to layers, that supports robust relational reasoning, in contrast to the dense model’s volatile and brittle organization. We hope this work will catalyze deeper dialogue between the MoE and interpretability communities and pave the way for training diagnostics that scale alongside model capacity.

\section{Method}

\subsection{MoE Architecture}
Mixture-of-Experts (MoE) models extend the standard Transformer architecture by introducing sparsely activated experts within certain layers. Each MoE layer contains $E$ independent feed-forward networks, known as experts, which share the same input dimension and structure but maintain separate parameters.

During inference, a gating network selects the top-$k$ experts for each input token based on their routing. The final output of an MoE layer is computed as a weighted sum of the selected expert outputs:
\begin{equation}
\text{MoE}(x) = \sum_{\mathcal{E} \in \text{Top-}k} \text{gate}_\mathcal{E}(x) \cdot \text{FFN}_\mathcal{E}(x),
\end{equation}
where $x$ is the residual stream input, $\text{gate}_\mathcal{E}(x)$ is the gating score for expert $\mathcal{E}$, and each expert FFN follows a two-layer structure with a nonlinearity:
\begin{equation}
\text{FFN}_\mathcal{E}(x) = W^{(2)}_\mathcal{E} \cdot \sigma(W^{(1)}_\mathcal{E} x).
\end{equation}

While dense Transformers process all tokens through the same parameters, MoE introduces conditional computation—only a small subset of experts is activated per token. This sparsity enables parameter-efficient scaling but also alters how information and knowledge are distributed across the network.

Knowledge attribution seeks to identify which internal components of a language model contribute to specific predictions or knowledge representations. The sparsity of MoE architectures presents challenges for such analysis, as activations are no longer uniform across layers or inputs. In this work, we study how knowledge is stored and routed within MoE models, introducing a neuron-level attribution method to trace the predictive contributions of individual components over time.

\subsection{Neuron Attribution in MoE Models}
We adopt a neuron attribution method originally proposed for dense models~\citep{yu2023neuron}, and extend it to MoE models~\citep{li-etal-2025-decoding}. This method allows us to attribute the model's prediction for a given token to individual neurons based on their effect on the output log-probability.  

\paragraph{Expert Neuron ($v^\mathcal{E}$) Definition.} In each selected expert, we define a \textbf{neuron} as a column vector $v$ in the output projection matrix $W^{(2)}_\mathcal{E}$ (also referred to as a subvalue). Its activation is controlled by an input-dependent scalar coefficient $m = \sigma(w^{(1)} \cdot x)$, where $w^{(1)}$ is a row in $W^{(1)}_\mathcal{E}$ (the corresponding subkey) and $x$ is the residual stream input. 

In MoE models with Top-$k$ routing, which is the standard approach where each token activates multiple experts~\citep{jiang2024mixtral,dai2024deepseekmoe,qwenmoe}, the final output added to the residual stream is the weighted sum over all selected experts:
\begin{equation}
\text{output}_v = \sum_{\mathcal{E} \in \text{Top-}k} \text{gate}_\mathcal{E}(x) \cdot m_\mathcal{E} \cdot v_\mathcal{E},
\end{equation}
where each $(m_\mathcal{E}, v_\mathcal{E})$ pair corresponds to a value neuron in expert $e$ selected for the input token. This formulation naturally extends neuron attribution to MoE models by incorporating the gated contributions of multiple expert neurons.

In other words, each FFN neuron corresponds to one column vector (subvalue) in the transformer’s output projection matrix rather than a single activation, and its strength is determined by the inner product between the input and its subkey, followed by a nonlinear transformation~\citep{yu2023neuron}.

\paragraph{Attention Neuron ($v^\mathcal{A}$) Definition.} For attention layers, we extend the notion of neuron attribution to \textbf{attention neurons}. Each multi-head attention layer uses per-head value and output projection matrices $W^V_j, W^O_j \in \mathbb{R}^{d \times d}$ for heads $j = 1, \dots, H$, applied to the value vectors $V$.
The attention output at position $i$ can be written as
\begin{equation}
\text{Attn}(x)_i = \sum_{j=1}^H \sum_{p=1}^T \alpha_{i,j,p} \cdot W^O_j\!\big(W^V_j h_p\big),
\end{equation}
where $\alpha_{i,j,p}$ is the attention weight computed via softmax
over the dot product between the query at position $i$ and the key at
position $p$ in head $j$.

We define the $k$-th column of $W^O_{j}$ as the $k$-th attention subvalue in head $j$, analogous to FFN subvalues. Its corresponding attention subkey is the $k$-th row of $W^V_{j}$. We refer to each (subvalue, subkey) pair as an attention neuron $v^\mathcal{A}$.

We attribute importance to each attention neuron by measuring the change in the log-probability of the target token when its contribution is added to the residual stream.

\textbf{Neuron Importance Scoring.} The importance of a neuron $v$ is defined by the increase in the log-probability of the target token $w$ when its output is added to the residual stream:
\begin{equation}
\text{I}(v) = \log p(w \mid x + \text{output}_v) - \log p(w \mid x).
\end{equation}
This formulation allows us to rank neurons by their predictive influence, and identify high-attribution neurons that encode semantic or syntactic information.

\section{Experiment Settings}
\subsection{Models} 
\begin{table*}[t]
\centering
\begin{tabular}{lcccccc}
\hline
Model Name & Layers & \begin{tabular}{@{}c@{}}FFN / Expert \\ Dimension\end{tabular} & \begin{tabular}{@{}c@{}}Number of \\ Experts\end{tabular} & \begin{tabular}{@{}c@{}}Routing \\ Strategy\end{tabular} & \begin{tabular}{@{}c@{}}Attention \\ Heads\end{tabular} & \begin{tabular}{@{}c@{}}Head \\ Dimension\end{tabular} \\
\hline
OLMo-7B & 32 & 11008 & -- & -- & 32 & 128 \\
OLMoE-1B-7B & 16 & 1024 & 64 & Top-8 & 16 & 128 \\
\hline
\end{tabular}
\caption{Model Configuration.}
\label{tab:model_config}
\end{table*}



We compare two open-source Transformer models released by the same lab, ensuring consistent implementation and comparable training pipelines: \textbf{OLMo-7B} (dense) and \textbf{OLMoE-1B-7B} (MoE). Their key architectural specifications are summarized in Table~\ref{tab:model_config}.

OLMo-7B~\citep{groeneveld2024olmo} serves as our dense baseline. It is a 7B-parameter decoder-only Transformer trained for 600K steps on 2.5T tokens from the Dolma corpus~\citep{soldaini-etal-2024-dolma}, which primarily consists of diverse web text and general-domain data.

As a sparse counterpart, OLMoE-1B-7B~\citep{muennighoff2024olmoe} implements a Mixture-of-Experts architecture with approximately 1.3B active parameters per token. Each of its 16 layers is an MoE block containing 64 independent experts, with a Top-8 routing mechanism. The model was trained for 1.2M steps on 5.0T tokens, following the same per-step token budget (4.2M tokens) as OLMo-7B. OLMoE-1B-7B uses the OLMoE-MIX corpus~\citep{muennighoff2024olmoeopenmixtureofexpertslanguage}—a blend of DCLM~\citep{li2024datacomp} and Dolma~\citep{soldaini-etal-2024-dolma} with greater weight on code/math—whereas OLMo-7B is trained on Dolma ($\geq$ 2T tokens). We acknowledge this composition difference as a potential confound and control for it in our analyses where applicable.


MoE models are typically trained for more steps to ensure sufficient expert utilization~\citep{li2025can} and to align with compute-optimal scaling principles~\citep{du2024revisiting,ludziejewski2025joint}. This extended training stabilizes the routing dynamics and allows experts to converge toward specialized functional roles. 


\subsection{Datasets}
To investigate knowledge attribution at the neuron level, we leverage a relational facts dataset. This dataset is a carefully selected subset from the comprehensive collection introduced by Hernandez et al.~\citep{hernandez2023linearity}, which was originally designed to probe for linear relational structures in large language models. Our subset is structured to provide a balanced evaluation across different knowledge domains.

It comprises a total of 906 subject-object examples, spanning 12 distinct relations from four principal categories: \textit{Linguistic}, \textit{Commonsense}, \textit{Factual}, and \textit{Bias}. Each data point consists of a subject-object pair (e.g., \textit{Paris, France}), a corresponding cloze-style prompt template (e.g., ``The capital of \{\} is the city of''), and the ground-truth object token. The specific relations included are:

\begin{itemize}
\item \textbf{Linguistic}: adjective\_antonym, word\_first\_letter, word\_last\_letter

\item \textbf{Commonsense}: object\_superclass, fruit\_inside\_color, work\_location

\item \textbf{Factual}: country\_language, country\_capital\_city

\item \textbf{Bias}: name\_religion, occupation\_age, occupation\_gender, name\_birthplace
\end{itemize}

This diverse selection of relations allows for a multi-faceted analysis, enabling us to examine how knowledge of varying types---from lexical patterns to world facts and learned social associations---is encoded and processed during pre-training in both MoE and dense architectures.

\subsection{Metrics}
We assess model performance using \textbf{HIT@10}, which evaluates the accuracy of relational knowledge prediction. Given a subject entity and a cloze-style prompt template (e.g., “The capital of \{\} is the city of”), the model is asked to predict the correct object entity (e.g., “Moscow”). The model generates a ranked list of candidate tokens, and HIT@10 measures the proportion of cases where the ground-truth object appears among the top-10 predictions.

\begin{equation}
\text{HIT@10} = \frac{1}{N} \sum_{i=1}^{N} \mathbb{I}[y_i \in \text{TopK}(\hat{\boldsymbol{y}}_i, 10)],
\end{equation}

where $N$ is the number of test examples, $y_i$ is the ground-truth token for the $i$-th example, and $\hat{\boldsymbol{y}}_i$ is the vector of predicted scores (e.g., logits) over the vocabulary. The indicator function $\mathbb{I}[\cdot]$ returns 1 if the condition is true and 0 otherwise.

To better evaluate the stability of neurons and layers throughout pre-training, we define four quantitative metrics:

\textbf{Top-1\% Set Stability ($J_{\mathrm{stab}}$).}
This metric quantifies the temporal consistency of the most important neurons across training checkpoints. 
At each step $t$, we identify the top-1\% most important neurons (based on absolute importance score $|I|$) and compute their Jaccard overlap with the next checkpoint:

\begin{equation}
J_{\mathrm{stab}} = \frac{1}{S-1} \sum_{t=1}^{S-1} 
\frac{| \mathcal{T}_t \cap \mathcal{T}_{t+1} |}{| \mathcal{T}_t \cup \mathcal{T}_{t+1} |},
\end{equation}

where $\mathcal{T}_t$ denotes the index set of the top-1\% neurons at step $t$, and $S$ is the total number of checkpoints. 
Higher $J_{\mathrm{stab}}$ values indicate stronger persistence of top neurons across time, reflecting more stable activation patterns.

\textbf{Positive-Gain Concentration ($R_t$).}
This metric measures how concentrated the positive importance updates are among the most reinforced neurons at each training step. 
Let $\Delta^{+}|I_{i,t}| = \max(0, |I_{i,t}| - |I_{i,t-1}|)$ denote the positive gain in importance for neuron $i$ at step $t$. 
Then the concentration ratio is defined as:

\begin{equation}
R_t = 
\frac{\sum_{i \in \mathcal{T}_t} \Delta^{+}|I_{i,t}|}
{\sum_{j \in \mathcal{N}} \Delta^{+}|I_{j,t}|},
\end{equation}

where $\mathcal{N}$ represents all neurons in the layer, and $\mathcal{T}_t$ corresponds to the top-1\% neurons ranked by $\Delta^{+}|I_{i,t}|$. 
A higher $R_t$ indicates that the learning signal is concentrated on a small subset of neurons, implying sparse but dominant functional adaptation.

\textbf{Layer-Distribution Consistency ($\rho_{\mathrm{avg}}$)} assesses the similarity of layer-wise importance score distributions across different training steps. A higher value indicates more stable and consistent usage patterns across the model's depth.

\begin{equation}
\rho_{\mathrm{avg}} = \frac{2}{S(S-1)} \sum_{1 \leq i < j \leq S} \text{Corr}(I_i, I_j),
\end{equation}

where $S$ is the total number of checkpoints, and $I_i = [I_{i,1}, \ldots, I_{i,L}]$ denotes the importance scores across $L$ layers at step $i$. The Pearson correlation coefficient is defined as:

\begin{equation}
\text{Corr}(I_i, I_j) = \frac{\sum_{l=1}^{L} (I_{i,l} - \bar{I}_i)(I_{j,l} - \bar{I}_j)}{\sigma_i \sigma_j},
\end{equation}

where $\bar{I}_i = \frac{1}{L} \sum_{l=1}^{L} I_{i,l}$ is the average importance scores across all layers at step $i$, and $\sigma_i^2 = \frac{1}{L} \sum_{l=1}^{L} (I_{i,l} - \bar{I}_i)^2$ is the corresponding variance.

\textbf{Cross-step Coefficient of Variation ($\sigma_{\mathrm{rel}}$)} measures the relative fluctuation amplitude of each layer's importance scores across their time series, where lower values indicate higher stability.

\begin{equation}
\sigma_{\mathrm{rel}} = \frac{1}{L} \sum_{l=1}^{L} \frac{\sigma_l}{|\mu_l|},
\end{equation}


where $\mu_l = \frac{1}{S} \sum_{s=1}^{S} I_{s,l}$ is the average importance scores across all steps for layer $l$, and $\sigma_l^2 = \frac{1}{S} \sum_{s=1}^{S} (I_{s,l} - \mu_l)^2$ is the corresponding variance.

\section{Neuron-Level Dynamics}
\label{tab:neuron_level}
In this section, we probe the micro-level evolution of model knowledge by tracking  FFN neurons in the MoE model ($v^\mathcal{E}$) and in the dense model ($v^\mathcal{F}$), as well as attention neurons ($v^\mathcal{A}$) with two complementary metrics introduced above ($J_{stab}$ and $R_t$). 

\subsection{MoE's Reinforced Core in FFN Neurons}
 
\begin{figure}[t]
\centering
\includegraphics[width=\columnwidth]{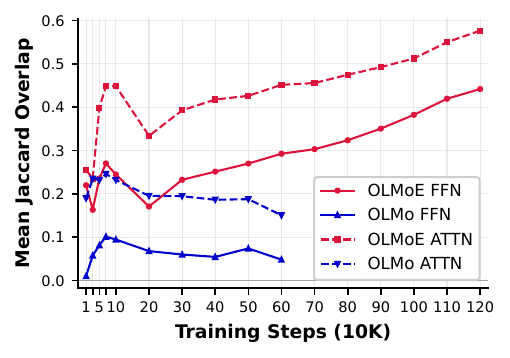}
\caption{Top-1\% FFN and ATTN neurons Jaccard overlap between consecutive checkpoints.}
\label{fig:neuron_jaccard}
\end{figure}

Figure~\ref{fig:neuron_jaccard} illustrates the trajectories of $\mathcal{J}_{\text{stab},t}$ across checkpoints, 
while Table~\ref{tab:neuron_stability_metrics} reports the averaged stability and concentration values. 
These results allow us to compare how dense and MoE models differ in the consistency and focus of their most important neurons over time.

As shown in Figure~\ref{fig:neuron_jaccard}, OLMoE exhibits markedly higher Top-1\% set stability and stronger concentration of positive updates. 
Specifically, OLMoE’s Jaccard curve follows a three-phase \textit{explore--refine--consolidate} trajectory: it rises to 27\% at 70K steps, dips to 17.1\% at 200K, and then climbs monotonically to 44.2\% by 1.2M steps. 
This dip reflects selective refinement rather than noise, since about 13\% of neurons survive across 100K--300K steps, compared to only 0.4\% under random. 
Table~\ref{tab:neuron_stability_metrics} further confirms this trend, showing that during consolidation the positive-gain concentration remains high 
($R_t \approx 45$--$52\%$), meaning that nearly half of all positive updates continually reinforce the same top-1\% neurons.

\begin{table}[t]
\centering
\small
\begin{tabular}{lccc}
\hline
\multirow{2}{*}{Model} & \multicolumn{2}{c}{\textbf{Mean $J_{stab}$ (\%)}} & \multirow{2}{*}{\textbf{Mean $R_t$ (\%)}} \\
\cline{2-3}
& 10K--70K & 100K--600K & \\
\hline
OLMo   & 6.3 & 6.1 & 26.4 \\
OLMoE  & 22.2 & 25.3 & 48.9 \\
\hline
\end{tabular}
\caption{Neuron-level stability metrics.}
\label{tab:neuron_stability_metrics}
\end{table}

In contrast, OLMo never develops a durable core. Its Jaccard overlap peaks briefly at 10.2\% around 70K steps, but then declines to 4.9\% by 600K. Likewise, $R_t$ is substantially lower (26.4\% on average), showing that positive updates are frequently shifted to newly promoted neurons rather than consolidating a stable set.

\subsection{MoE's Sustained Stability in ATTN Neurons}
To complement our findings on FFN neurons ($v^\mathcal{E}$ and $v^\mathcal{F}$), we apply the same two-metric analysis to ATTN neurons $v^\mathcal{A}$, aiming to assess whether neurons in attention heads exhibit similar stability during pre-training.

There is a significant difference in the stability of the core neuron sets. As shown in Figure~\ref{fig:neuron_jaccard}, the Jaccard overlap of the two models follows distinctly different trajectories. OLMoE again follows the \textit{explore--refine--consolidate} trajectory. Specifically, its Jaccard overlap rises rapidly to 44.9\% at 100K steps after an initial exploration, dips to 33.4\% at 200K steps. We further calculated the overlap between the top-1\% neurons at 100K and 300K steps to be 26.73\%. This value, being lower than the consecutive overlap from 100K--200K, confirms the existence of the "refine" phase, yet it also shows that roughly a quarter of the core neurons were preserved through this adjustment and formed the foundation for the subsequent stable set. To verify this value, we also computed the overlap for a randomly selected 1\% of neurons between the same steps, which averaged only 0.005. However, this adjustment paves the way for a more optimal convergence: thereafter, the overlap enters a sustained and robust upward trajectory, ultimately reaching 0.577 at 1200K steps.


Importantly, the consolidation dynamics, measured by the proportion of importance score from the top-1\% neurons ($R_t$) relative to the total positive importance gain across the layer, perfectly complements the observed Jaccard trajectory. Specifically, this importance score ratio steadily climbs from an initial 21.5\%, stabilizing in a plateau of 33\%--34\% after 200K steps. These trends are also clearly reflected in the metrics summarized in Table~\ref{tab:neuron_stability_metrics}, which shows that OLMoE exhibits substantially higher neuron-level stability and average importance ratio than OLMo. Together, these results demonstrate that OLMoE is actively identifying and consistently reinforcing a stable core subset of $v^\mathcal{A}$ throughout training.

By comparison, OLMo fails to consolidate a stable set of attention neurons. Its Jaccard overlap peaks at 24.6\% around 70K steps but then steadily declines to 15.1\% by 600K. This downward trajectory reflects continual replacement and high churn, with important neurons frequently discarded rather than stabilized.

Both FFN and ATTN layers in OLMoE converge toward reinforced cores of neurons that remain consistently important throughout training, whereas OLMo exhibits only transient overlaps without durable consolidation.

\section{Layer-Level Stability During Pre-training}
In this section, we now ask whether this consolidation propagates upward to the layer scale. We assess this by tracking the stability of layer-wise importance profiles using two complementary metrics ($\rho_{\mathrm{avg}}$ and $\sigma_{\mathrm{rel}}$).

\subsection{Early Consolidation of FFN Layers in MoE}
\begin{figure}[t]
\centering
\includegraphics[width=\columnwidth]{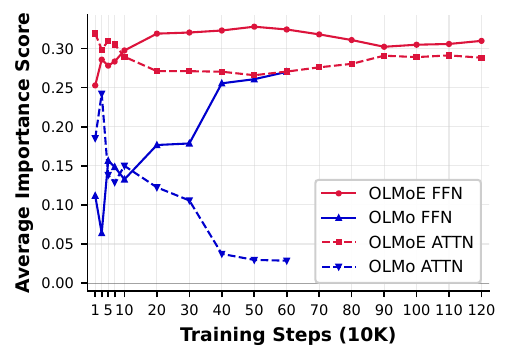}
\caption{Mean FFN and ATTN importance scores across all layers over training steps. OLMoE shows smoother and earlier stabilization compared to OLMo.}
\label{fig:all_avg}
\end{figure}

As shown in Figure~\ref{fig:all_avg}, the average FFN importance for OLMoE rapidly increases and plateaus early in training, reaching 0.298 by 100K steps and peaking at 0.328 by 500K steps with minimal fluctuation. By contrast, OLMo’s FFN curve is unstable, dipping to 0.064 at 30K steps before climbing irregularly to 0.270 by 600K steps. 

This difference in stability is further quantified in Table~\ref{tab:stability_metrics_all}. OLMoE achieves a near-perfect $\rho_{\mathrm{avg}}$ of 0.97 between 10K and 600K steps, indicating that its layer-wise importance profile locks in early. OLMoE's $\sigma_{\mathrm{rel}}$ is an order of magnitude lower than OLMo's (\textbf{0.37 vs. 5.01}), showing that its layer-wise importance profile locks in early and drifts minimally across steps. Conversely, OLMo’s $\rho_{\mathrm{avg}}$ is only 0.54, reflecting weaker convergence and greater variability. 

\begin{figure}[t]
\centering
\includegraphics[width=\columnwidth]{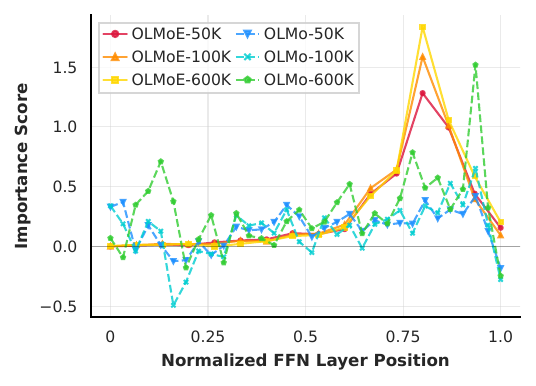}
\caption{Snapshots of FFN layer importance distribution at different training steps.}
\label{fig:ffn_layer}
\end{figure}

Figure~\ref{fig:ffn_layer} further illustrates this contrast: 
OLMoE’s FFN importance distribution is nearly identical between 50K and 600K steps, whereas OLMo continues to exhibit noticeable shifts across layers.

Together, these findings suggest that the MoE architecture promotes an efficient and early consolidation of FFN layer utility, a distinction from the protracted optimization process in its dense counterpart.

\begin{table}[t]
\centering
\small
\begin{tabular}{c cc cc}
\hline
\multirow{2}{*}{\textbf{Model}} & \multicolumn{2}{c}{\textbf{FFN Metrics}} & \multicolumn{2}{c}{\textbf{ATTN Metrics}} \\[0.3ex]
\cline{2-5}
& \textbf{$\rho_{\mathrm{avg}}$} & \textbf{$\sigma_{\mathrm{rel}}$} & \textbf{$\rho_{\mathrm{avg}}$} & \textbf{$\sigma_{\mathrm{rel}}$} \\[0.3ex]
\hline
OLMo (10K--600K) & 0.54 & 5.01 & 0.49 & 8.64 \\[0.5ex]
OLMoE (10K--600K) & 0.97 & 0.37 & 0.97 & 0.49 \\[0.5ex]
OLMoE (10K--1.2M) & 0.97 & 0.47 & 0.95 & 0.76 \\[0.5ex]
\hline
\end{tabular}
\caption{Comparison of FFN and ATTN stability metrics. $\rho_{\mathrm{avg}}$: Layer-Distribution Consistency; $\sigma_{\mathrm{rel}}$: Cross-step Coefficient of Variation.}
\label{tab:stability_metrics_all}
\end{table}

\subsection{Sustained Stability of ATTN Layers in MoE}

\begin{figure}[t]
\centering
\includegraphics[width=\columnwidth]{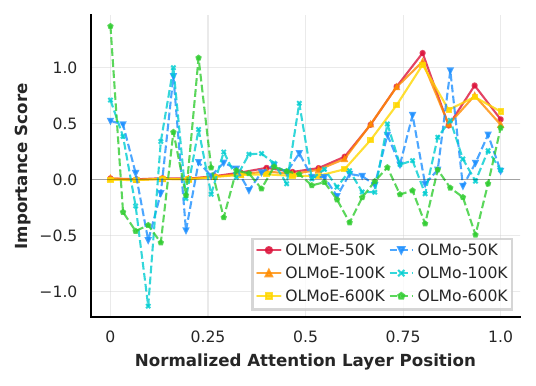}
\caption{Snapshots of ATTN layer importance distribution at different training steps.}
\label{fig:attn_layer}
\end{figure}

We next investigate whether this stabilization is unique to FFN layers or a global property of the MoE architecture by analyzing the ATTN layers. The dynamics of ATTN layers present a different but equally revealing pattern. The two models present a complementary pattern between their FFN and ATTN layer importance. As illustrated in Figure~\ref{fig:all_avg}, OLMoE's average ATTN importance remains high and stable throughout training, starting at 0.319 and settling into a narrow band around 0.290. In sharp contrast, OLMo's ATTN importance undergoes a monotonic decline. This suggests that as the dense model's FFN layers consolidate their utility, its attention heads become increasingly redundant.

\begin{figure*}[t]
\centering
\includegraphics[width=\textwidth]{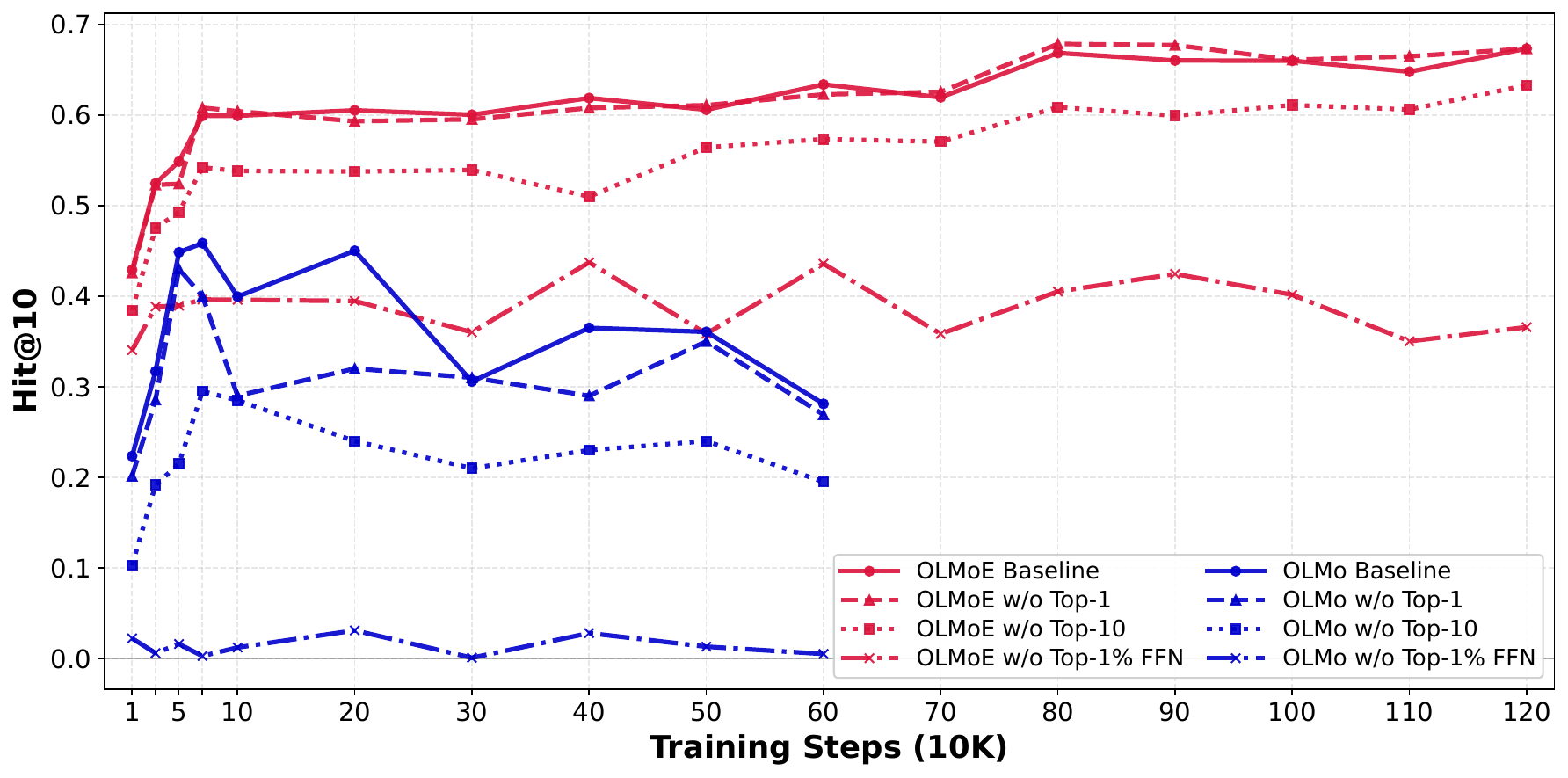}
\caption{Performance of the model under ablation interventions (Top-1 head, Top-10 heads and Top-1\% FFN neurons) measured by the HIT@10 metric.}
\label{fig:ablation}
\end{figure*}

The stability metrics in Table~\ref{tab:stability_metrics_all} confirm this divergence. OLMoE achieves a high $\rho_{\mathrm{avg}}$ (0.97) and a low $\sigma_{\mathrm{rel}}$ of 0.49, demonstrating a stable attention layer profile that is visually supported by the stable importance distributions in Figure~\ref{fig:attn_layer}. OLMo, with a $\rho_{\mathrm{avg}}$ (\textbf{0.49}) and a $\sigma_{\mathrm{rel}}$ of \textbf{8.64}, shows the opposite: its attention layer hierarchy is continuously reconfigured with high volatility throughout training, which we have already discussed from a micro perspective---the neuron level.

In summary, the layer-level stability observed in this section can be regarded as the macro-level projection of the phenomena described in the previous section, namely the sustained reinforcement of the top-1\% neurons. This suggests that, during training, MoE propagates stability from a micro-level core toward an early stabilization of the layer-wise importance profile, thereby providing the structural foundation for the inference-time robustness analyzed in the following section.

\section{Functional Robustness: Causal Intervention}

Our preceding analysis showed that the MoE develops a stable computational structure from neuron level to layer level, whereas the dense remains volatile across training. This raises a functional question: does early structural stabilization translate into robustness at inference time, or merely into a brittle concentration of knowledge? If this stability reflects distributed knowledge storage rather than brittle concentration, then ablating a small, attribution-selected set of important components should incur only minor drops—especially for MoE. 

To test this, we performed causal interventions by masking the most important neurons and attention heads to measure the performance drop on the HIT@10 metric, which evaluates the accuracy of relational knowledge prediction.





Ablating the Top-1\% FFN neurons reveals a MoE–dense contrast. OLMoE exhibits a moderate performance reduction of 35.5\% as shown in Table~\ref{tab:ablation} and Figure~\ref{fig:ablation}, consistent with a distributed knowledge backbone. OLMo nearly collapses, with a 96.1\% mean drop, and the performance approaches zero under the same intervention.

\begin{table}[t]
\centering
{\small
\begin{tabular}{lccc}
\hline
\textbf{Model} & 
\textbf{Top-1 head} & 
\textbf{Top-10 heads} & 
\begin{tabular}[c]{@{}c@{}}\textbf{Top-1\%}\\[-2pt]\textbf{FFN neurons}\end{tabular} \\
\hline
OLMoE & 0.06\%  & 9.44\%  & 35.47\% \\
OLMo  & 16.46\% & 50.43\% & 96.19\% \\
\hline
\end{tabular}
}
\caption{Performance Drop (\%) from Masking.}
\label{tab:ablation}
\end{table}

In the Top-10 heads study, OLMoE demonstrates remarkable resilience similar to the neuron ablation, causing only a minor 9.44\% average performance drop. As shown in Figure~\ref{fig:ablation}, the performance drop is not only substantially smaller than in OLMo, but this drop progressively shrinks as training advances. This stands in stark contrast to OLMo, where the performance drop is large, volatile and averages a catastrophic 50.43\%, showing no clear trend towards greater resilience. This is further clarified by the Top-1 head ablation. Ablating the single most important head in OLMo causes a performance drop (16.46\%), whereas the same intervention has a negligible effect on OLMoE. This indicates that OLMoE learns resilience by distributing knowledge across components, rather than concentrating it in a few critical ones.

These findings provide evidence that a model's architecture is a decisive factor in shaping its function. We show that OLMoE’s learning process, characterized by the stable reinforcement of a broad neuron set, is closely associated with its robust, distributed knowledge. This stable internal structure, formed early and reinforced consistently, makes the model resilient to perturbations. OLMo’s continual churn is the very mechanism that underlies its fragility: by concentrating knowledge in an ever-changing set of transient, brittle components, the dense architecture creates a vulnerable system. These results support the view that stability fosters robustness, while volatility leads to fragility.

\section{Related Work}

\paragraph{Mixture of Experts}
Mixture-of-experts (MoE) models were originally proposed as gated ensembles of specialized predictors, where a learned router softly partitions the input space and combines expert outputs~\citep{jacobs1991adaptive,jordan1994hierarchical}. Modern MoE architectures expand model capacity to address computational cost via conditional computation in large neural networks~\citep{bengio2013estimating,cho2014exponentially,shazeer2017outrageously,lepikhin2020gshard,fedus2022switch}. \citet{lepikhin2020gshard} introduced large-scale sparse activation via softmax Top-2 routing and an auxiliary load-balancing loss. Subsequent refinements, such as the router \emph{z-loss}~\citep{zoph2022st}, further stabilized optimization. Recent MoE language models typically adopt Top-$k$ ($k>1$) routing to improve expert utilization—e.g., Mixtral~\citep{jiang2024mixtral}, OLMoE~\citep{muennighoff2024olmoe}, and Qwen-MoE~\citep{qwenmoe}—and enhance specialization through shared or routed experts~\citep{shazeer2017outrageously,dai2024deepseekmoe,li2025dynamic}.

\paragraph{Mechanism Interpretability.}
Mechanism interpretability can be viewed along two complementary axes: (A) post-hoc analyses that recover internal mechanisms of trained models, and (B) training-time analyses that trace how such mechanisms emerge. For (A), neuron-level attribution methods quantify importance: Integrated Gradients (IG)~\citep{sundararajan2017axiomatic} and its neuron-level extensions~\citep{shrikumar2018computationally} support applications such as “knowledge neurons” and pruning for compression~\citep{dai2021knowledge,yvinec2022singe,ali2025detecting}; Layer-wise Relevance Propagation (LRP) is adapted to transformers with principled propagation across attention/residual paths and improved conservation/robustness~\citep{chefer2021transformer,ali2022xai}; Log-Probability Increase (LPI) scores neuron importance via log-likelihood deltas under neuron insertion/removal~\citep{yu2023neuron}, enabling analyses in multi-hop reasoning and bias and extending to MoE expert specialization in post-hoc settings~\citep{yu2025back,li-etal-2025-decoding}. Complementary to these, the circuits line decomposes transformer into verifiable subgraphs of attention heads and MLP components~\citep{conmy2023towards,bhaskar2024finding}. For (B), recent studies track when mechanisms form during pre-training: linear probes over hundreds of checkpoints show “trust” dimensions becoming separable early, while circuits-based analyses reveal evolving internal subgraphs that support knowledge acquisition~\citep{qian2024towards,ou-etal-2025-llms}. Most approaches focus on the trained dense model. This raises a question: how do architectural differences influence knowledge acquisition during pre-training?

\section{Conclusion}
In this paper, we conducted a comparison of knowledge acquisition dynamics in MoE and dense models across their pre-training, enabled by a gated attribution method that tracks neuron importance beyond traditional post-hoc analysis. Our findings reveal that MoE architectures form a low-entropy backbone of consistently reinforced neurons, which leads to an early consolidation of their importance profiles and, in turn, underpins their functional robustness. This resilience, stemming from more distributed knowledge storage, contrasts with the greater brittleness and knowledge concentration in the dense model. These phenomena collectively demonstrate that architectural sparsity is not merely a computational shortcut but also acts as a useful inductive bias that fosters stable and robust learning, offering insights for the design of stability-aware expert pruning and explore–then–stabilize routing strategies in future.


\section{Acknowledgments}
This work was supported by the National Natural Science Foundation of China (Grant No.62506318); Guangdong Provincial Department of Education Project (Grant No.2024KQNCX028); CAAI-Ant Group Research Fund; Scientific Research Projects for the Higher-educational Institutions (Grant No.2024312096), Education Bureau of Guangzhou Municipality; Guangzhou-HKUST(GZ) Joint Funding Program (Grant No.2025A03J3957), Education Bureau of Guangzhou Municipality.

\bibliography{aaai2026}

\end{document}